%% file: main.tex
\documentclass[10pt,twocolumn,letterpaper]{article} 

\usepackage{avss}
\usepackage{times}
\usepackage{epsfig}
\usepackage{graphicx}
\usepackage{amsmath}
\usepackage{amssymb}
\usepackage{multirow}
\usepackage{arydshln}
\usepackage{subcaption}

\avssfinalcopy 

\def\avssPaperID{****} 

\begin{document}

\title{SOP$^2$: Transfer Learning with Scene-Oriented Prompt Pool on 3D Object Detection}


\input{sections/0_authors}

\maketitle

\input{sections/0_abstract}

\input{sections/1_introduction}
\input{sections/2_related_work}

\input{sections/3_proposed_method}

\input{sections/4_experiments}

\input{sections/5_conclusion}


{\small
\bibliographystyle{ieee}
\bibliography{main}
}

\end{document}


\title{SOP$^2$: Transfer Learning with Scene-Oriented Prompt Pool on 3D Object Detection}


\maketitle

\input{sections/6_supp}


%% file: sections/0_authors.tex
\author{
Ching-Hung Cheng$^{1}$, Hsiu-Fu Wu$^{2}$, Bing-Chen Wu$^{1}$, Khanh-Phong Bui$^{1}$,Van-Tin Luu$^{1}$, \\Ching-Chun Huang$^{1}$\thanks{Corresponding author} \\
$^{1}$National Yang Ming Chiao Tung University, Hsinchu, Taiwan \\
$^{2}$Internet of Things Laboratory, Chunghwa Telecom Laboratories, Taoyuan, Taiwan \\
{\tt\small \{hong3924.cs10, evan20010126.cs12, phongbk.ee13, tinery.ee12, chingchun\}@nycu.edu.tw} \\
{\tt\small q104769424@cht.com.tw}
}

%% file: sections/0_abstract.tex
\begin{abstract}
With the rise of Large Language Models (LLMs) such as GPT-3, these models exhibit strong generalization capabilities. Through transfer learning techniques such as fine-tuning and prompt tuning, they can be adapted to various downstream tasks with minimal parameter adjustments. This approach is particularly common in the field of Natural Language Processing (NLP). This paper aims to explore the effectiveness of common prompt tuning methods in 3D object detection. We investigate whether a model trained on the large-scale Waymo dataset can serve as a foundation model and adapt to other scenarios within the 3D object detection field. This paper sequentially examines the impact of prompt tokens and prompt generators, and further proposes a Scene-Oriented Prompt Pool (\textbf{SOP$^2$}). We demonstrate the effectiveness of prompt pools in 3D object detection, with the goal of inspiring future researchers to delve deeper into the potential of prompts in the 3D field.
\vspace{-0.5cm}

\end{abstract}

%% file: sections/1_introduction.tex
\section{Introduction}

3D object detection is a crucial task in computer vision, enabling applications such as autonomous driving, robotics, augmented reality, and urban planning. Large-scale datasets like Waymo~\cite{sun2020scalability} and KITTI~\cite{geiger2012we} have significantly advanced this field, yet models trained on one dataset often fail to generalize well across different domains due to domain gaps. These gaps arise from variations in sensor configurations, environmental conditions, and data collection methodologies, ultimately leading to degraded performance when models are deployed in new environments. Addressing this challenge is essential for developing robust and adaptable 3D object detection systems.

One common strategy to bridge the domain gap is Unsupervised Domain Adaptation (UDA), which aims to improve model generalization without requiring labeled target domain data. UDA enables models to learn shared feature representations between source and target domains, enhancing cross-domain performance. However, due to the complexity of designing effective training strategies and loss functions, UDA models often struggle to extract meaningful features from unlabeled target data, resulting in suboptimal adaptation. Fig.~\ref{fig_teaser}a illustrates how UDA improves generalization by leveraging target domain data.

Recent advancements in Natural Language Processing (NLP) have introduced Large Language Models (LLMs) like GPT-3~\cite{floridi2020gpt}, which learn from massive datasets and demonstrate strong generalization capabilities. Instead of fine-tuning all parameters, researchers have developed Parameter-Efficient Fine-Tuning (PEFT) methods, such as Low-Rank Adaptation (LoRA)~\cite{hu2021lora} and Prompt Tuning~\cite{lester2021power}, to reduce computational costs while preserving adaptability. LoRA, as shown in Fig.~\ref{fig_teaser}b, approximates linear layers with low-rank matrices to minimize trainable parameters. Meanwhile, Prompt Tuning (Fig.~\ref{fig_teaser}c) introduces task-specific prompts into the model input, optimizing performance with minimal adjustments.

An emerging approach, Prompt Pools, enhances prompt tuning by storing multiple prompt tokens and dynamically selecting the most relevant ones during inference, as illustrated in Fig.~\ref{fig_teaser}d. Inspired by these advancements, this paper investigates the role of prompt tuning in 3D object detection, exploring whether a model trained on the Waymo dataset can serve as a foundation model for other domains. We systematically analyze prompt tokens and prompt generators, ultimately proposing the Scene-Oriented Prompt Pool (\textbf{SOP$^2$}) to improve domain adaptation. Our work demonstrates the effectiveness of prompt-based adaptation in 3D object detection, providing insights for future research on leveraging LLM-inspired techniques in 3D vision.

\definecolor{darkgreen}{RGB}{42, 150, 29}
\definecolor{darkred}{RGB}{241, 71, 58}
\begin{figure}
    \begin{center}
    \includegraphics[width=0.4\textwidth]{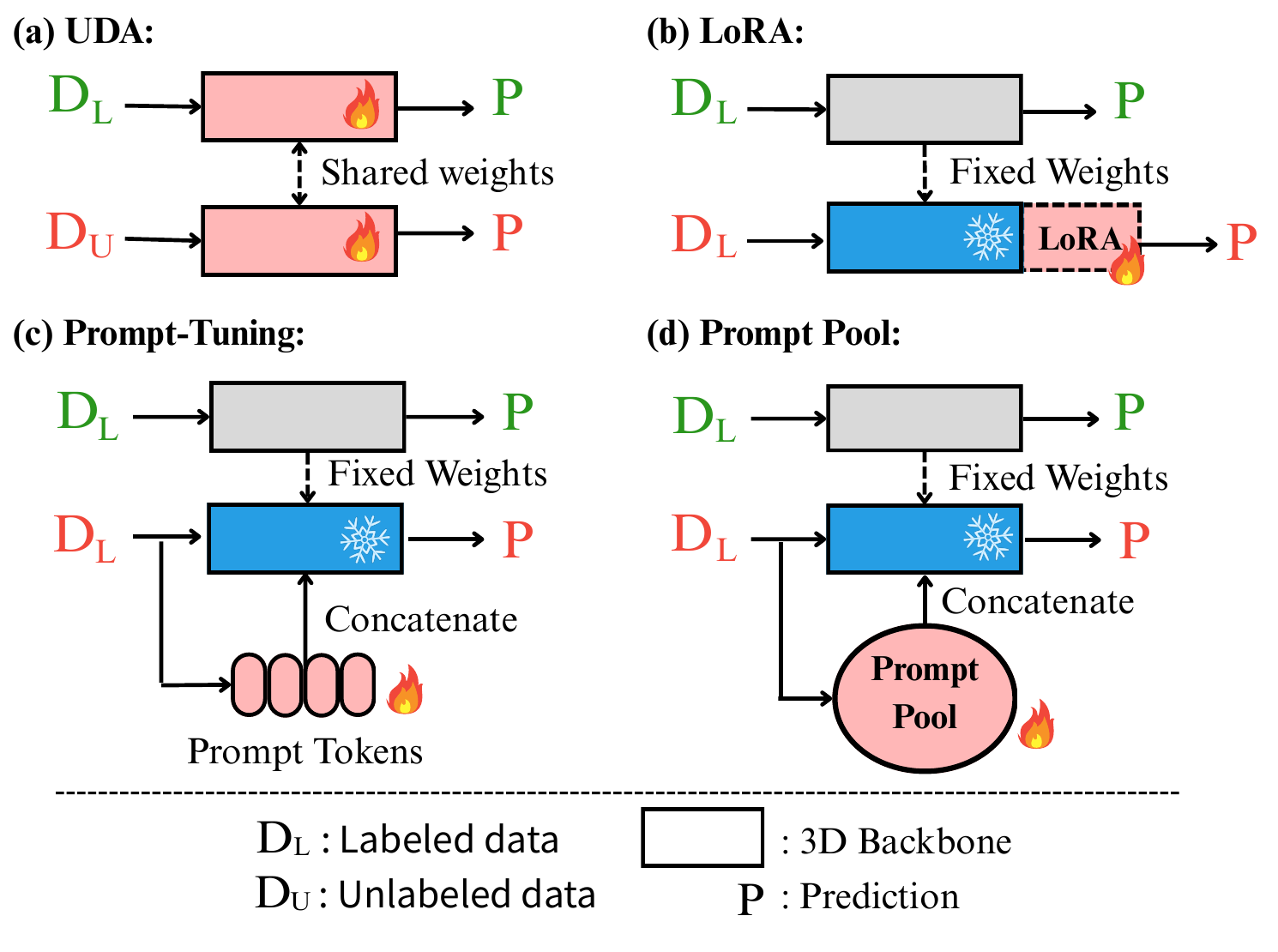}
    \end{center}
    \vspace{-0.5cm}

    \caption{Comparison of various transfer learning methods: (a) Unsupervised Domain Adaptation, (b) Low-Rank Adaptation, (c) Prompt Tuning and (d) Prompt Pool. \textcolor{darkgreen}{Green text}  represents the source domain, while \textcolor{darkred}{red text} represents the target domain.}
    \vspace{-0.5cm}

\label{fig_teaser}
\end{figure}

%% file: sections/2_related_work.tex
\begin{figure*}[htb]
  \centering
  \begin{subfigure}{0.495\textwidth}
    \includegraphics[width=\linewidth]{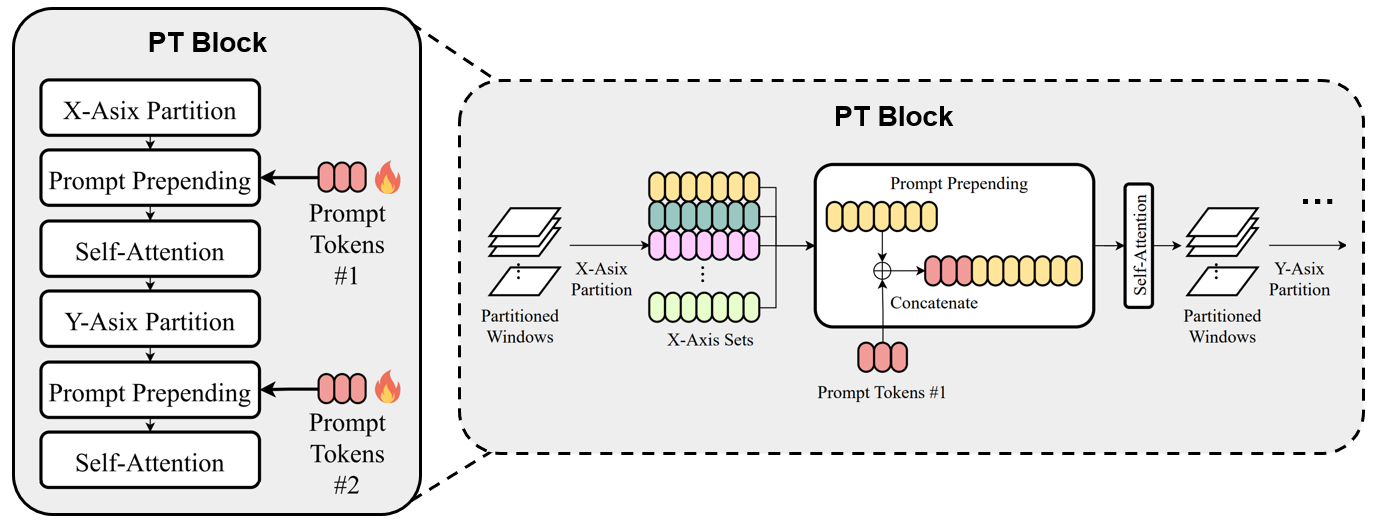}
    \vspace{-0.8cm}
    \caption{Prompt Token} 
    \label{fig_PT_PG:a}
  \end{subfigure}
  \hfill
  \begin{subfigure}{0.495\textwidth}
    \includegraphics[width=\linewidth]{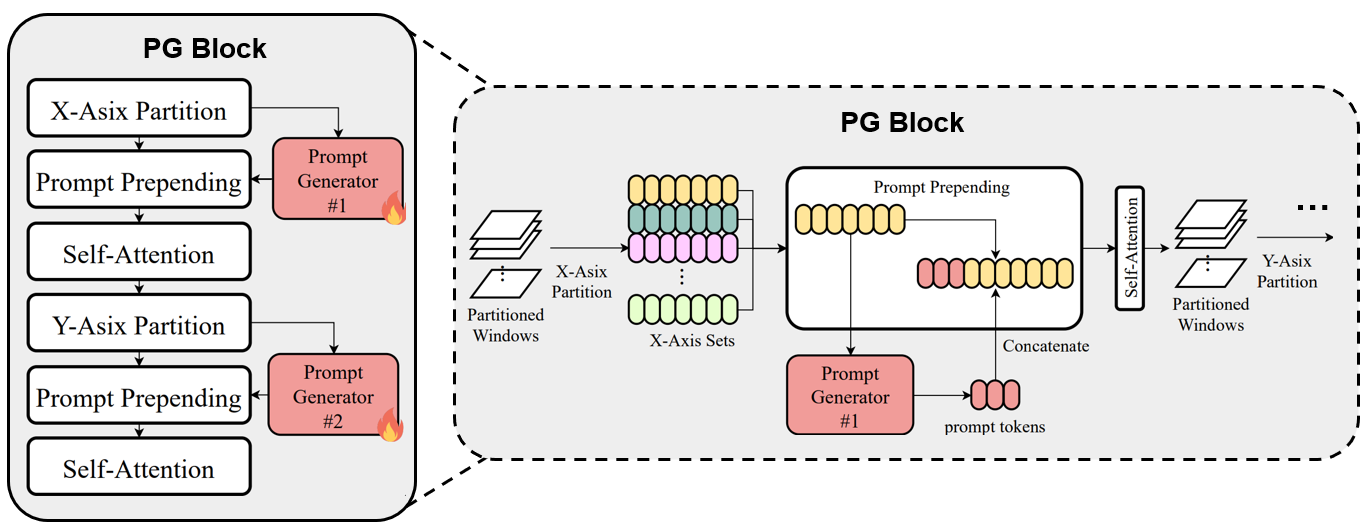}
    \vspace{-0.8cm}
    \caption{Prompt Generator} 
    \label{fig_PT_PG:b}
  \end{subfigure}
\vspace{-0.2cm}

\caption{The architecture of (a) PT block: Adding prompt token to $\textbf{\textit{S}}_j$ and (b) PG block: Adding prompt generator to $\textbf{\textit{S}}_j$.}
\label{fig_PT_PG}
\vspace{-0.5cm}

\end{figure*}
\section{Related Work}


\subsection{Parameter-Efficient Fine-Tuning}


Recent research in fine-tuning has focused on Parameter-Efficient Fine-Tuning (PEFT) to adapt pretrained models more effectively, as traditional fine-tuning often incurs high training costs. Two prominent techniques, Prompt-Tuning and LoRA, have gained attention for their efficiency in adapting pretrained models to specific tasks. Prompt-Tuning fine-tunes models by introducing task-specific prompts, enabling adaptation with minimal supervision. It guides the model to focus on relevant information for the target task, improving performance while preserving pretrained benefits. On the other hand, LoRA addresses the computational challenges of fine-tuning large models by using low-rank approximations of weight matrices, reducing the parameter space and enabling more efficient fine-tuning with faster convergence, making it suitable for resource-constrained environments.

\subsection{Prompt-Tuning in CV}
While Prompt-Tuning traditionally involves providing task-specific prompts to guide the fine-tuning of pretrained language models, visual prompt tuning (VPT) \cite{jia2022visual}, extends this idea to the realm of visual data. Instead of text prompts, VPT utilizes visual cues or soft prompts to guide the adaptation of pretrained vision models for specific tasks. This approach offers a promising avenue for leveraging pretrained representations in CV tasks while allowing for task-specific adaptation with minimal labeled data. Dynamic visual prompt tuning (DVPT) \cite{ruan2023dynamic} further extends this approach by dynamically generating prompts based on the input data. Unlike VPT, where prompts remain fixed throughout the training process, prompts in DVPT adapt based on the characteristics of the current input, allowing for more flexible and context-aware model behavior.

\subsection{Transformer for 3D Perception}


Transformers \cite{vaswani2017attention} have gained significant traction in computer vision, particularly for 3D perception tasks. Votr \cite{mao2021voxel}, a Transformer-based model, excels at extracting 3D features by capturing global dependencies between voxels in point clouds. However, its dense architecture incurs high computational costs, especially when scaling to large 3D datasets. To mitigate this, Window Attention techniques \cite{liu2021swin} have emerged, offering a more efficient approach to 3D perception. Building on this, SST \cite{9878875} introduces a locality-aware spatial attention mechanism that dramatically reduces computational overhead, while SWFormer \cite{sun2022swformer} further refines this by incorporating a sliding window-based mechanism to strike a balance between accuracy and efficiency. More recently, DSVT \cite{wang2023dsvt} has taken sparse transformer models to new heights by dynamically generating sparse voxel representations, enabling efficient modeling of global dependencies while significantly lowering computational costs. This approach has pushed the boundaries of Transformer-based 3D perception, showcasing its vast potential.

%% file: sections/3_proposed_method.tex
\section{Proposed Method}

\subsection{Prompt Friendly Framework}\label{subsec:friendly}

DSVT is well-suited for prompt-based 3D perception due to its efficient sparse data handling and dynamic attention mechanism, reducing computational overhead and improving scalability. By dynamically adjusting attention per prompt, it enhances contextual understanding across scene components. We adopt a pillar-based DSVT framework, where the input point cloud is processed through a voxel feature encoding (VFE) module \cite{yin2021center}, followed by max-pooling along the voxel dimension. The scene is partitioned into windows, with nonzero voxels evenly distributed along the X/Y axes to form non-overlapping sets. Each DSVT block applies two set partitions: one along the X-axis and one along the Y-axis. The $j$-th partition is defined as: 
\begin{equation}
    \textit{\textbf{S}}_j = \{S_i\}_{i=1}^N, \ j = 1,...,Total \ \# \ of \ Partitions,
    \label{eq:sets}
\end{equation}

where $S_i \in \mathbb{R}^{n_s \times C}$ represents a set of $n_s$ voxels with feature dimension $C$, and $N$ denotes the total number of sets, depending on voxel sparsity. The full partition is expressed as $\textit{\textbf{S}}_j \in \mathbb{R}^{N \times n_s \times C}$. To enable information exchange within each set, multi-head self-attention (MHSA) is applied.

We consider $N$ as the batch size, treating voxels as tokens. By leveraging multi-scale window partitioning and directional set partitioning, DSVT facilitates efficient token interaction, enhancing 3D scene understanding.

\subsection{Scene Analysis}
\label{Scene Analysis}
Unlike previous task-oriented prompt-tuning approaches such as \cite{lester2021power,vu2021spot}, which focus on adapting pre-trained models for downstream tasks, we address the shift from 2D image classification to 3D point cloud object detection. This shift has increased data volume, and existing pre-trained models are less powerful and generalizable than large LLMs. We propose a scene-oriented prompt approach, where prompt tokens focus on learning domain-specific information to capture the unique characteristics of the target domain. We further analyze the scene using t-SNE \cite{van2008visualizing} on partitioned sets $\textbf{\textit{S}}_j$, observing distinct distribution patterns, as shown in Fig. \ref{fig_tsne:a}. This suggests the need for independent prompt tokens for each partition, rather than a single generic prompt, ensuring better alignment with the specific characteristics of each data subset and enhancing the model's ability to capture domain-specific features.

\subsection{Prompt Token}
\label{Prompt Token}
In the VPT framework, a group of prompt tokens are concatenated to the embedding patches prior to the model input. During the fine-tuning process, only the prompt tokens are updated, while the backbone model is kept frozen. By adopting this approach, the prompt tokens are able to learn the information relevant to the downstream tasks, while the backbone model retains the original source domain knowledge.
Taking inspiration from VPT, in Fig.~\ref{fig_PT_PG:a}, for each set partition $\textbf{\textit{S}}_j$, we assign a collection of prompt tokens $PT_j$ as follows:
\begin{equation}
    PT_j, \ j = 1,...,Total \ \# \ of \ Partitions,
    \label{eq:PT}
\end{equation}
where $PT_j \in \mathbb{R}^{n_T \times C}$ is the prompt tokens corresponds to $\textbf{\textit{S}}_j$, $n_T$ is the number of prompt. Then we cooperate prompt and scene as follows:
\begin{equation}
    \textit{\textbf{S}}_j^* = \left\{\left[ PT_j;S_i\right]\right\}_{i=1}^N = \left\{ S_i^*\right\}_{i=1}^N,
    \label{eq:prompted_Sj_PT}
\end{equation}
where ; denoted concatenation along the dimension of token numbers. we cooperate the prompt token $PT_j$ with each set $S_i$ in $j$-th set partition to obtain the prompted sets $S_i^*$, where $S_i^* \in \mathbb{R}^{(n_T+n_s) \times C}$, $n_T$ is the prompt number of $PT_j$. Then, we conduct the MHSA as mentioned in Section~\ref{subsec:friendly} on the prompted set partition $\textbf{\textit{S}}_j^*$ 

In Section \ref{main_results}, we found that simply using a prompt token has limited effectiveness.


\begin{figure*}[htb]
    \centering
    \includegraphics[width=0.9\textwidth]{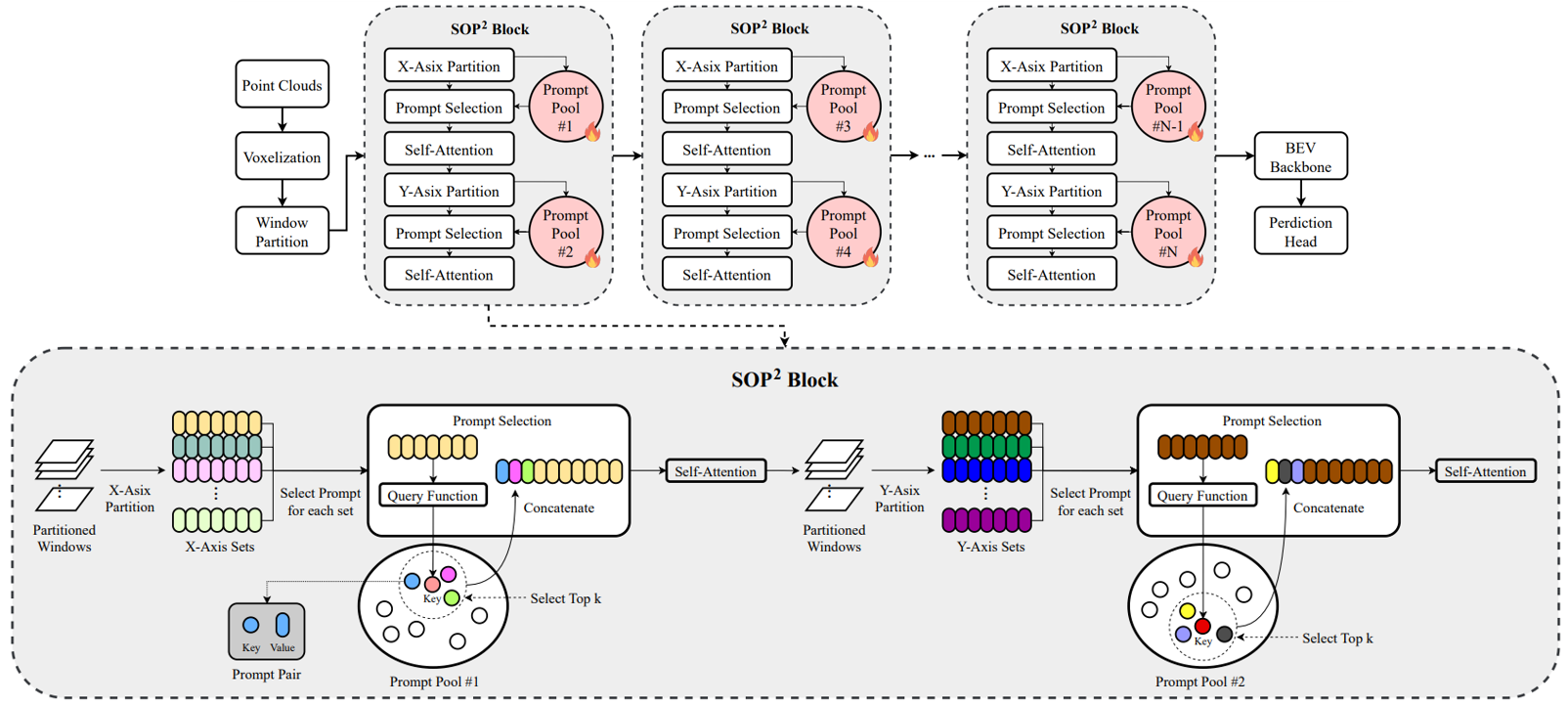}
    \caption{The overall architecture of our proposed SOP$^2$. The upper part shows the overall pipeline, while the lower part illustrates a single SOP$^2$ block. Each set partition is assigned a corresponding prompt pool, allowing the set to select suitable prompt tokens from the prompt pool.}
    \label{fig_pp_dsvt}
    \vspace{-0.5cm}

\end{figure*}

\subsection{Prompt Generator}
\label{Prompt Generator}
Since providing a simple prompt for each set may not be sufficient for the model, we further draw inspiration from DVPT to enhance prediction performance. Instead of a simple prompt token, a prompt generator is employed to generate a prompt token of the same size as the set token, ensuring that each prompt can refer to the current input scene. As shown in Fig.~\ref{fig_PT_PG:b}, we design a prompt generator $f^G: \mathbb{R}^{n_s \times C} \rightarrow \mathbb{R}^{n_G \times C}$ to dynamically produce prompt tokens corresponding to $\textbf{\textit{S}}_j$. We denote the $j$-th prompt generator as:
\begin{equation}
    f_j^G, \ j = 1,...,Total \ \# \ of \ Partitions
    \label{eq:PG}
\end{equation}
The sets $S_i$ in $j$-th set partition are fed into $f_j^G$ to get the set-wise dynamic prompts $P_i^G$, where $P_i^G \in \mathbb{R}^{n_G \times C}$, $n_G$ is the prompt number of $P_i^G$. Then we cooperate the set-wise prompts and scene as follows:
\begin{equation}
    \textit{\textbf{S}}_j^* = 
    \left\{\left[ f_j^G(S_i);S_i \right]\right\}_{i=1}^N = 
    \left\{\left[ P_i^G;S_i \right]\right\}_{i=1}^N = 
    \left\{ S_i^*\right\}_{i=1}^N,
    \label{eq:prompted_Sj_PG}
\end{equation}
where the prompted sets $S_i^* \in \mathbb{R}^{(n_G+n_s) \times C}$. In Section \ref{main_results}, we found that even with the addition of a prompt generator to dynamically generate prompts, it may still be insufficient for 3D scenes.

\subsection{Scene-Oriented Prompt Pool (SOP$^2$)}
\label{SOPP}
After the aforementioned attempts of prompt token and prompt generator, it was observed that using prompts generated based on the scene is more effective than general prompts. We further draw on the concept of the prompt pool from L2P \cite{wang2022learning} by storing prompt tokens in a prompt pool, allowing the current scene to select the most suitable prompt token for use. Building upon the previous attempts and the concept of L2P. As shown in Fig.~\ref{fig_pp_dsvt}, we conducted a prompt pool $PP_j$ with the corresponding $\textbf{\textit{S}}_j$, we denoted the prompt pool as follows:
\begin{equation}
    PP_j = \left\{ (k_m, P_m) \right\}_{m=1}^M, \ M = size \ of \ prompt \ pool,
    \label{eq:PP}
\end{equation}
where $k_m$ and $P_m$ are key-value pair, $k_m \in \mathbb{R}^{C}$ and $P_m \in \mathbb{R}^{n_P \times C}$, $n_P$ is the length of prompt. We aim to dynamically select the most suitable  prompts from the prompt pool. and cooperate these prompts with $S_i$ which is set-wise. The selection function of prompt pool is as follows:
\begin{equation}
    f^s(PP_j, f^q(S_i)) = P_i^P, \ where \ P_i^P \in \mathbb{R}^{K \times n_P \times C}
    \label{eq:fs}
\end{equation}

First, we use a query function $f^q:\mathbb{R}^{n_s \times C} \rightarrow \mathbb{R}^{C}$ to project the input set to the same dimension of key, the output of $f^q$ is the query key, then $f^s$ will score the similarity of all $k_m$ in prompt pool $PP_j$ and select the top-$K$ similar keys with the query key. Finally, collect the corresponding value of the top-$K$ key to construct the set-wise prompt $P_i^P$. Then we integrate the set-wise prompts with the scene to obtain the prompted set $S_i^* \in \mathbb{R}^{(K \times n_p+n_s) \times C}$ as follows:

\begin{equation}
\label{eq:prompted_Sj_PP}
  \begin{aligned}
    \textit{\textbf{S}}_j^* &= 
    \left\{\left[ f^s(PP_j,f^g(S_i));S_i \right]\right\}_{i=1}^N \\
    &= \left\{\left[ P_i^P;S_i \right]\right\}_{i=1}^N = 
    \left\{ S_i^*\right\}_{i=1}^N,
    \end{aligned}
\end{equation}


%% file: sections/4_experiments.tex
\section{Experiments Results}
\subsection{Experimental Setup}
\noindent\textbf{Dataset Preparation.} The experimental section uses the DSVT pillar version with pre-trained weights from the Waymo dataset to evaluate prompt tuning on the KITTI dataset. The KITTI dataset includes 3,712 \textit{train}, 3,769 \textit{val}, and 7,518 \textit{test} samples, commonly used for 3D object detection. In comparison, the Waymo Open Dataset is much larger, with 158,361 \textit{train}, 40,077 \textit{val}, and 40,832 \textit{test} samples, collected across diverse locations and weather conditions, providing a more generalized dataset for model development.
\vspace{0.5em} 

\noindent\textbf{Evaluation Metrics.} We use the official KITTI evaluation metric to assess our method, based on mean Average Precision (mAP) with a rotated Intersection over Union (IoU) threshold. For cars, the threshold is 0.7, and for pedestrians and cyclists, it is 0.5. The mAP is calculated on the validation set with 40 recall positions, providing a comprehensive measure of object detection accuracy, including localization and classification.

\noindent\textbf{Implementation Details.} We utilize the same learning rate scheme as \cite{yin2021center} and strictly follow the DSVT setup. The backbone consists of four DSVT blocks, each containing two set partitions (X and Y Axis). We adopt a grid size of (0.32m, 0.32m, 6m) and hybrid window sizes of (12, 12, 1) and (24, 24, 1). The maximum number of voxels assigned to each set, $n_s$, is set to 36. The voxel feature map is downsampled using standard max-pooling along the Z-Axis. All attention modules utilize 192 input channels. In Section \ref{Prompt Token}, $n_T$, the number of prompt tokens $PT_j$, is set to 1. In Section \ref{Prompt Generator}, the prompt generator $f^G$ is a 4-layer MLP, and the number of prompts $n_G$ is set to 1. In Section \ref{SOPP}, the query function $f^q$ is a max pooling function, and $f^s$ uses cosine similarity to score the query and prompt key.

\input{table/main_results}

\input{table/pool_position}
\input{table/runtime}

\begin{figure}[t]
  \centering
  \begin{subfigure}{0.23\textwidth}
    \includegraphics[width=\linewidth]{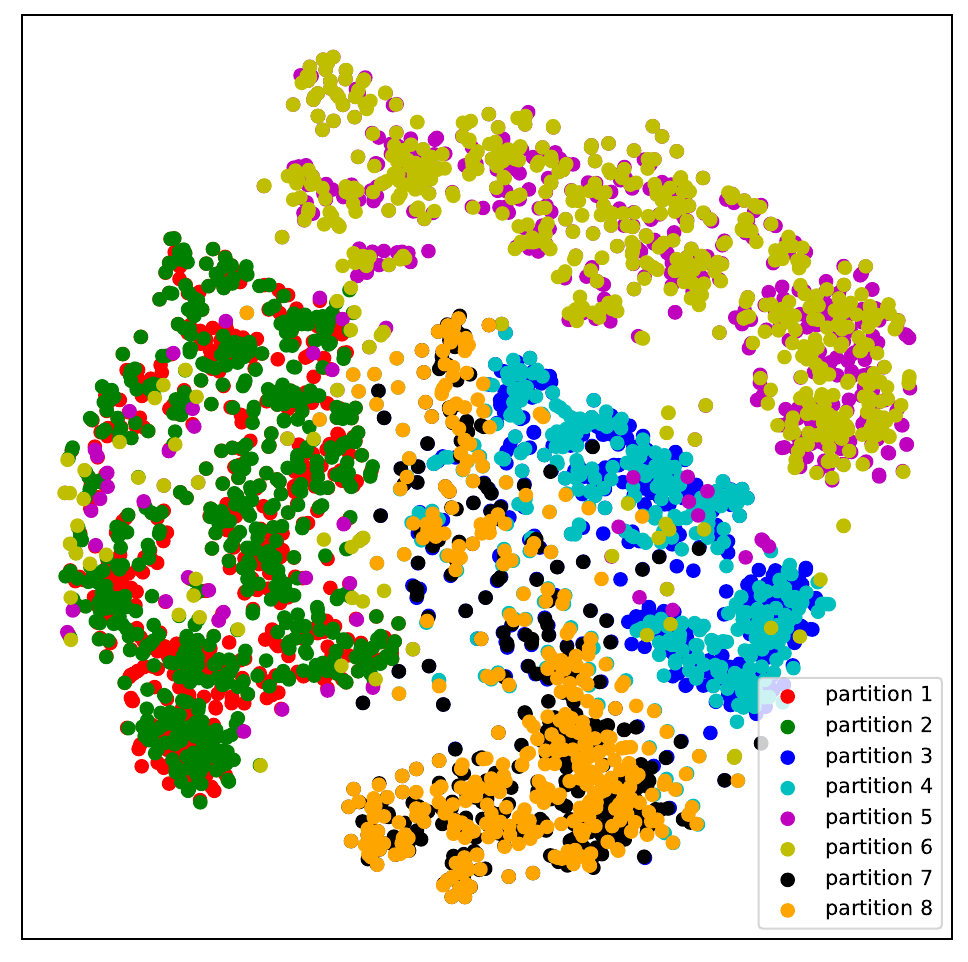}
    \vspace{-0.6cm}

    \caption{Set Partitions $\textbf{\textit{S}}_j$}
    \label{fig_tsne:a}
  \end{subfigure}
  \begin{subfigure}{0.23\textwidth}
    \includegraphics[width=0.99\linewidth]{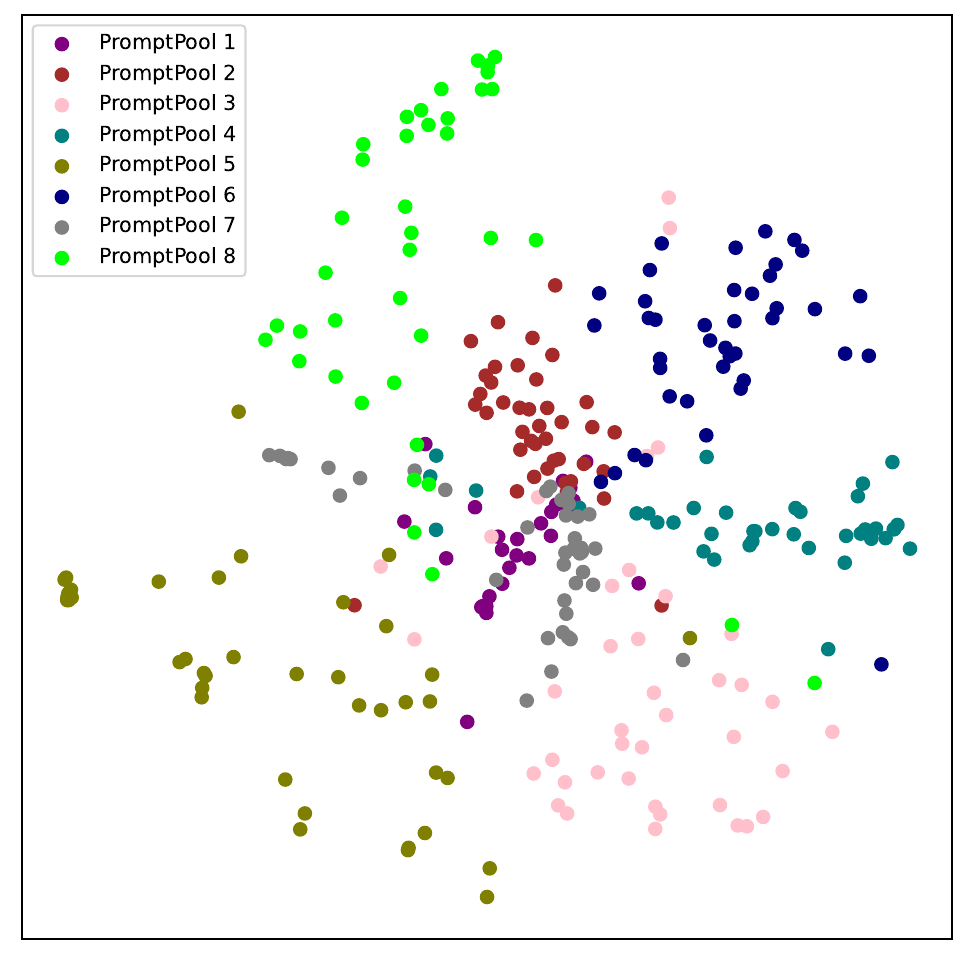}
    \vspace{-0.6cm}

    \caption{Prompt Pools $PP_j$} \label{fig_tsne:b}
  \end{subfigure}
\vspace{-0.3cm}

\caption{For the visual representation of t-SNE. (a) Distribution of different set partitions $\textbf{\textit{S}}_j$, where different colors represent different partitions, and (b) Distribution of different prompt pools $PP_j$ corresponding to set partitions $\textbf{\textit{S}}_j$, where different colors represent different prompt pools.}
\label{fig_tsne}
\end{figure}

\begin{figure}[t]
    \begin{center}
    \includegraphics[width=0.23\textwidth]{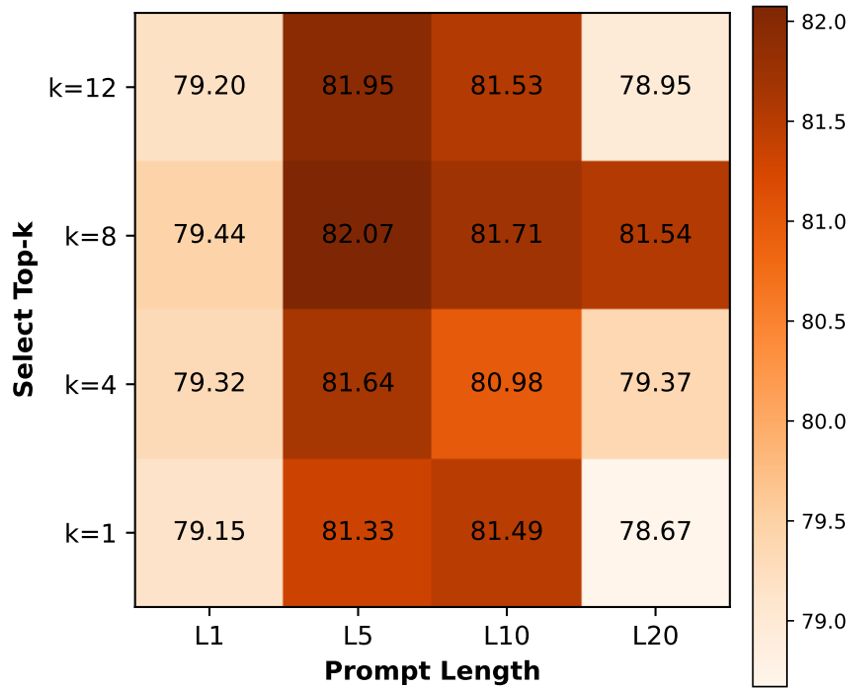}
    \end{center}
    \vspace{-0.5cm}

    \caption{Comparison of 3D mAP with 40 recall positions for Prompt Length $n_P$ and Select top-\textit{K} when the prompt pool size M = 40.}
   \label{fig_heat_map}
    \vspace{-0.5cm}

\end{figure}


\subsection{Main Results}
\label{main_results}
As shown in Table \ref{table:main}, we first train the DSVT model from scratch on the KITTI dataset as the baseline. We then fine-tune the model using Waymo pre-trained weights and explore two approaches: \textbf{Full Fine-Tuning} (updating all weights) and \textbf{Head Fine-Tuning} (updating only the Prediction Head to avoid overfitting). We also investigate \textbf{BitFit} (updating only the backbone's bias terms) and \textbf{LoRA}, a PEFT method using low-rank approximation to reduce trainable parameters in the transformer layers. Additionally, we test a \textbf{Prompt Token} method, where prompt tokens store domain-specific information for each set. Despite its simplicity, this method underperforms the From Scratch baseline. To improve upon this, we propose the \textbf{Prompt Generator} method (Section \ref{Prompt Generator}), which dynamically generates prompts based on the scene. This method outperforms the Prompt Token approach by 1.66\%, 4.3\%, and 1.52\% for Cars, Pedestrians, and Cyclists, respectively, under moderate conditions.

We also introduce \textbf{SOP$^2$} (Section \ref{SOPP}), inspired by the L2P prompt pool concept. SOP$^2$ uses prompt pools for each set partition, enabling the model to select the most suitable prompts for the current scene. SOP$^2$ outperforms the Prompt Generator by 1.63\%, 1.96\%, and 1.80\% for the three moderate classes, and the Prompt Token method by 3.29\%, 6.26\%, and 3.32\%. Compared to LoRA, SOP$^2$ shows a significant advantage of 1.17\%, 9.25\%, and 1.28\%. Finally, when combining SOP$^2$ with LoRA, the results demonstrate a 0.1\% improvement in Car class performance compared to SOP$^2$ alone, indicating effective synergy between SOP$^2$ and PEFT methods.

Regarding parameter count, we report the number of trainable parameters, excluding frozen weights, including the 0.45M parameters in the detection head. SOP$^2$ uses only 0.82M parameters (9\% of Full Fine-Tuning), which is 0.35M more than LoRA but achieves over 1\% higher accuracy, showing an excellent trade-off between parameter count and performance.


\subsection{Effectiveness of Prompt Pool}

\noindent\textbf{Prompt Pool Position.} Section \ref{Scene Analysis} discusses t-SNE analysis on set partitions $\textbf{\textit{S}}_j$, revealing distinct distributions, as shown in Figure~\ref{fig_tsne:a}. To capture partition-specific information, we assign a prompt pool to each partition. Further t-SNE analysis of the prompt pool values in Figure~\ref{fig_tsne:b} confirms that each pool encodes distinct information. Table \ref{table:partition} shows that odd-numbered partitions benefit more from the prompt pool, suggesting that X-axis partitions contribute more effectively than Y-axis partitions.

\begin{figure}[t]
    \begin{center}
    \includegraphics[width=0.4\textwidth]{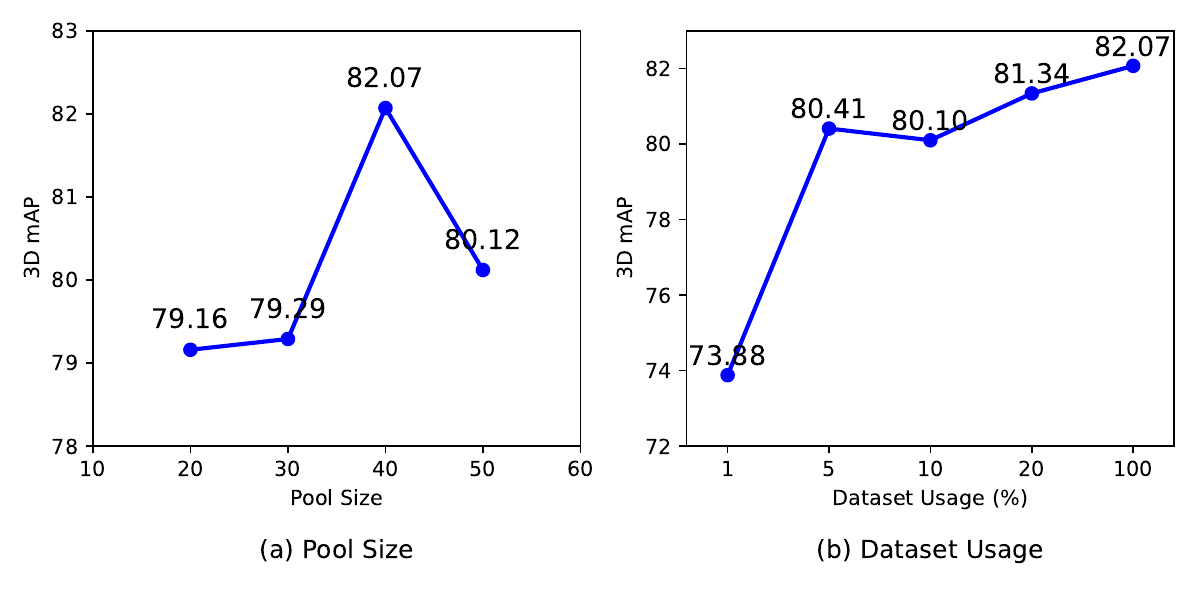}
    \end{center}
    \vspace{-0.7cm}
    \caption{Comparison of 3D mAP with 40 recall positions: (a) Effect of different pool sizes M, and (b) Effect of the quantity of the target dataset (KITTI) used for training SOP$^2$.}
   \label{fig_line}
    \vspace{-0.5cm}

\end{figure}

\noindent\textbf{Hyperparameters of Prompt Pool.} We first analyze the prompt pool size, as shown in Fig. \ref{fig_line}a, with $M$ ranging from 20 to 50. The optimal performance is achieved at $M = 40$, suggesting that an appropriately sized pool is crucial for effective information accommodation. Next, we examine the relationship between prompt length $n_P$ and top-$K$ selection for $M = 40$, as illustrated in Fig.~\ref{fig_heat_map}. The best performance occurs at $n_P = 5$ and $K = 8$, with performance degrading for larger or smaller values, potentially due to underfitting or overfitting influenced by the target domain size.

\noindent\textbf{Complexity}. Table \ref{tab:runtime} compares the baseline (DSVT) with our model, showing that adding the prompt pool causes only a slight increase in complexity. Inference time was measured on an Intel(R) Xeon(R) Silver 4210R CPU @ 2.40GHz and an NVIDIA RTX 3090 GPU.

\noindent\textbf{Target Domain Dataset Usage.} We analyzed the effect of the target domain data size (KITTI) on SOP$^2$ performance, as shown in Fig. \ref{fig_line}b. Using 1\%, 5\%, 10\%, 20\%, and 100\% of the KITTI training set (3712 samples), the results show that with only 1\% of the data, the 3D mAP is around 73.6, indicating underfitting. As the data usage increases, performance improves, with a sharp increase to 80.41 at 5\%. At 10\%, the accuracy slightly drops to 80.10, but further increases to 81.34 at 20\% and reaches 82.07 at 100\%, demonstrating continuous performance enhancement.
\vspace{-0.1cm}

%% file: table/main_results.tex
\begin{table}[tb] \scriptsize
    \centering
    \renewcommand{\arraystretch}{1.2}
    \setlength{\tabcolsep}{3pt} 
    \caption{Performance comparison on the KITTI $val$ set. The results are evaluated by the mean Average Precision with 40 recall positions. Best results among all methods except Full Fine-tune and From Scratch are \textbf{bolded}.}
    \label{table:main}
    \resizebox{0.475\textwidth}{!}{%
    \begin{tabular}{l|ccc|ccc|ccc|c}
        \hline
        \multirow{2}{*}{Method} & \multicolumn{3}{c|}{Car} & \multicolumn{3}{c|}{Pedestrian} & \multicolumn{3}{c|}{Cyclist} & Trainable\\
                                 & Easy & Mod. & Hard      & Easy & Mod. & Hard        & Easy & Mod. & Hard      & Params\\
        \hline
        From Scratch             & 90.08 & 81.88 & 80.22 & 53.47 & 48.54 & 45.20 & 92.68 & 70.91 & 66.65 & 8.65M\\
        \hline
        Head Fine-tune           & 88.51 & 78.87 & 77.45 & 57.81 & 51.69 & 47.57 & 82.57 & 63.57 & 60.47 & 0.45M\\
        Full Fine-tune           & 89.85 & 83.32 & 81.19 & 60.44 & 55.81 & 52.04 & 93.81 & 72.25 & 68.23 & 8.65M\\            
        \hline
        BitFit~\cite{zaken2021bitfit} & 88.14 & 79.94 & 79.28 & 59.97 & 54.70 & 50.13 & 86.00 & 66.67 & 63.49 & 0.47M\\
        LoRA~\cite{hu2021lora}        & 89.00 & 80.90 & 79.45 & 56.66 & 51.87 & 49.11 & 86.82 & 66.96 & 63.42 & 0.47M\\
        \hline
        Prompt Token~\cite{jia2022visual} & 88.55 & 78.78 & 77.65 & 61.95 & 54.86 & 50.10 & 83.60 & 64.92 & 61.50 & 0.45M\\
        Prompt Generator~\cite{ruan2023dynamic} & 89.03 & 80.44 & 79.13 & 65.99 & 59.16 & 54.71 & 83.71 & 66.44 & 63.10 & 1.64M\\
        \hline
        \textbf{SOP$^2$ (Ours)}        & 90.72 & 82.07 & 80.34 & \textbf{67.24} & 61.12 & \textbf{56.43} & \textbf{86.69} & \textbf{68.24} & \textbf{64.56} & 0.82M\\
        \textbf{SOP$^2$ (Ours)}$+$LoRA & \textbf{90.79} & \textbf{82.17} & \textbf{80.38} & 66.34 & \textbf{61.15} & 56.20 & 84.18 & 67.90 & 63.98 & 0.84M\\
        \hline
    \end{tabular}
    }
    \vspace{-0.3cm}

\end{table}

%% file: table/pool_position.tex
\begin{table}[tb]
    \centering
    \caption{Analysis of adding a prompt pool $PP_j$ to different corresponding set partitions $\textbf{\textit{S}}_j$.}
    \label{table:partition}
    \resizebox{0.35\textwidth}{!}{   
    \begin{tabular}{cccccccc|c}  
        \hline
        $Pa_1$&$Pa_2$&$Pa_3$& $Pa_4$&$P_5$& $Pa_6$& $Pa_7$&$Pa_8$&3D mAP\\
        \hline
        \checkmark & & & & & & & &80.34 \\
        &\checkmark & & & & & & &79.77 \\
        & &\checkmark & & & & & &81.21 \\
        & & &\checkmark & & & & &80.65 \\
        & & & &\checkmark & & & &81.54 \\
        & & & & &\checkmark & & &80.83 \\
        & & & & & &\checkmark & &80.64 \\
        & & & & & & &\checkmark &80.60 \\
        \hdashline
        \checkmark &\checkmark &\checkmark &\checkmark &\checkmark &\checkmark &\checkmark &\checkmark &82.07 \\
        \hline
    \end{tabular}
    }
    \vspace{-0.2cm}

\end{table}

%% file: table/runtime.tex
\begin{table}[ht]
\centering
\caption{Comparison of inference speed, memory usage, parameter count, and FLOPs.}
\resizebox{0.35\textwidth}{!}{%
\begin{tabular}{|l|c|c|c|c|}
\hline
\textbf{Method} & \textbf{Inference Time} & \textbf{Memory Peak} & \textbf{Params} & \textbf{FLOPs} \\
\hline
DSVT & 88.74 ms & 2951.47 MB & 8.65 M & 1460.66 G \\
Ours & 96.64 ms & 2953.23 MB & 9.02 M & 1464.68 G \\
\hline
\end{tabular}
}

\label{tab:runtime}
\vspace{-0.4cm}

\end{table}

%% file: sections/5_conclusion.tex
\section{Conclusion}
This paper investigates the feasibility of prompts in 3D perception, extending prompt tuning from LLMs and 2D vision to the 3D domain. Unlike previous \textbf{Task-Oriented} prompt tokens, we focus on \textbf{Scene-Oriented} prompts that encode scene-specific information. We study Prompt Tokens and Generators, proposing \textbf{SOP$^2$}, which dynamically selects the best prompt from multiple pools for each scene. \textbf{SOP$^2$} outperforms common fine-tuning and PETF methods, paving the way for future 3D prompt research.

%% file: sections/6_supp.tex
\section{Supplementary Material}

\begin{figure*}[h]
    \centering
    \includegraphics[width=\textwidth]{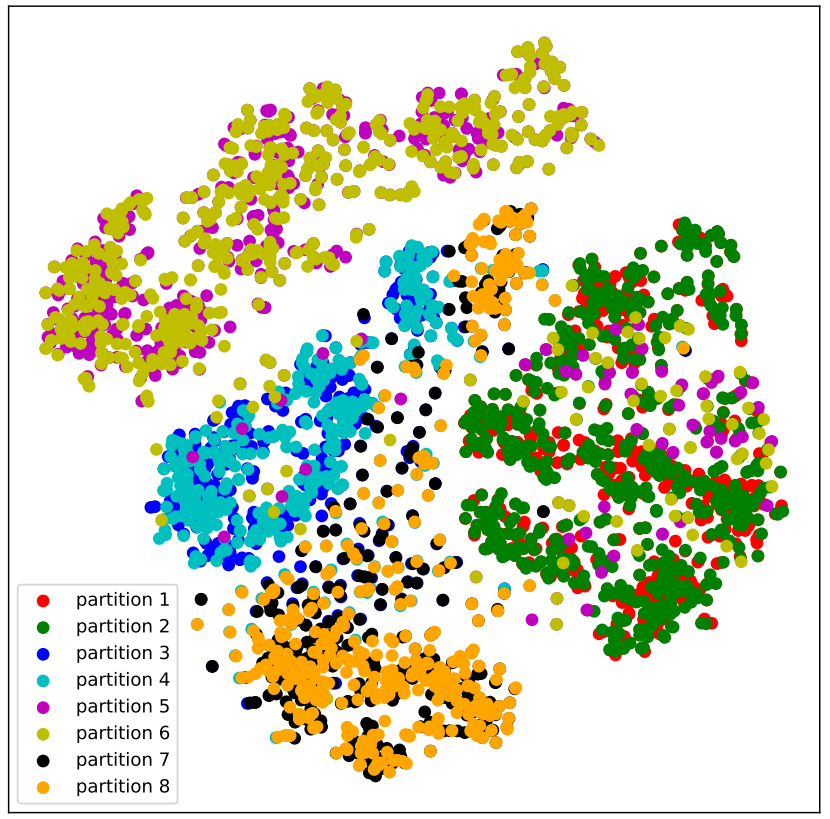}
    \caption{t-SNE visualization}
    \label{fig1}
\end{figure*}

\noindent As a supplement to Scene Analysis, we observed distinct distribution patterns for different set partitions $\textbf{\textit{S}}_j$. To further illustrate these observations, we provide additional t-SNE visualizations for deeper exploration, as shown in Figures~\ref{fig1}, \ref{fig2}, and \ref{fig3}. 

For the supplement to the Prompt Pool Position analysis, we performed t-SNE on the values within the prompt pool. The results revealed that different prompt pools correspond to distinct pieces of information. To facilitate further observation, we provide additional t-SNE visualizations in Figures~\ref{fig4}, \ref{fig5}, and \ref{fig6}.

\begin{figure*}[h]
    \centering
    \includegraphics[width=\textwidth]{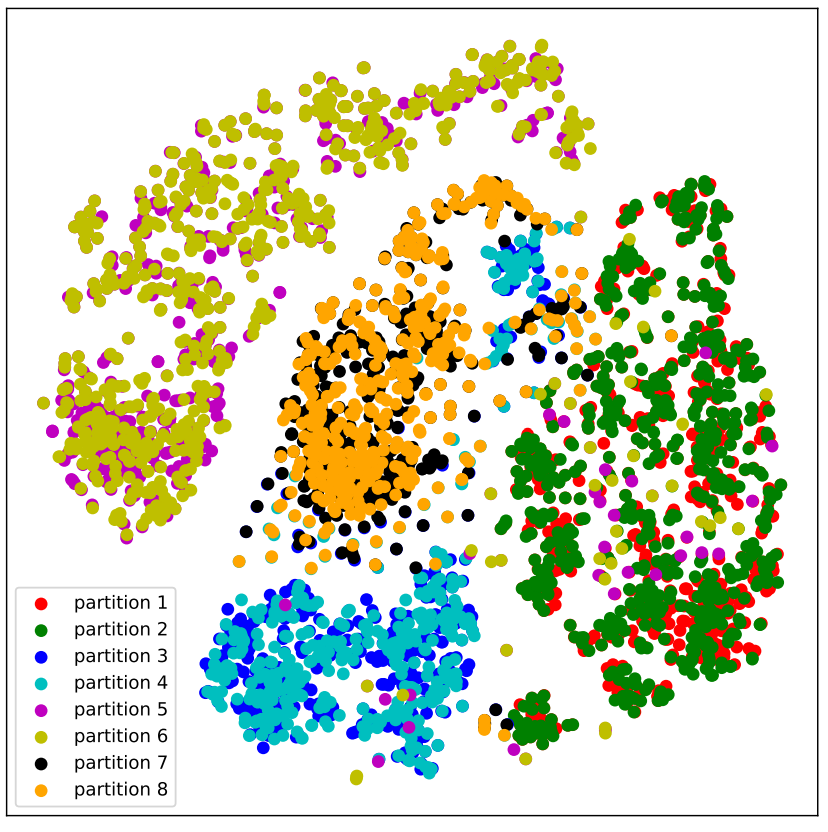}
    \caption{t-SNE visualization}

    \label{fig2}
\end{figure*}
\begin{figure*}[h]
    \centering
    \includegraphics[width=\textwidth]{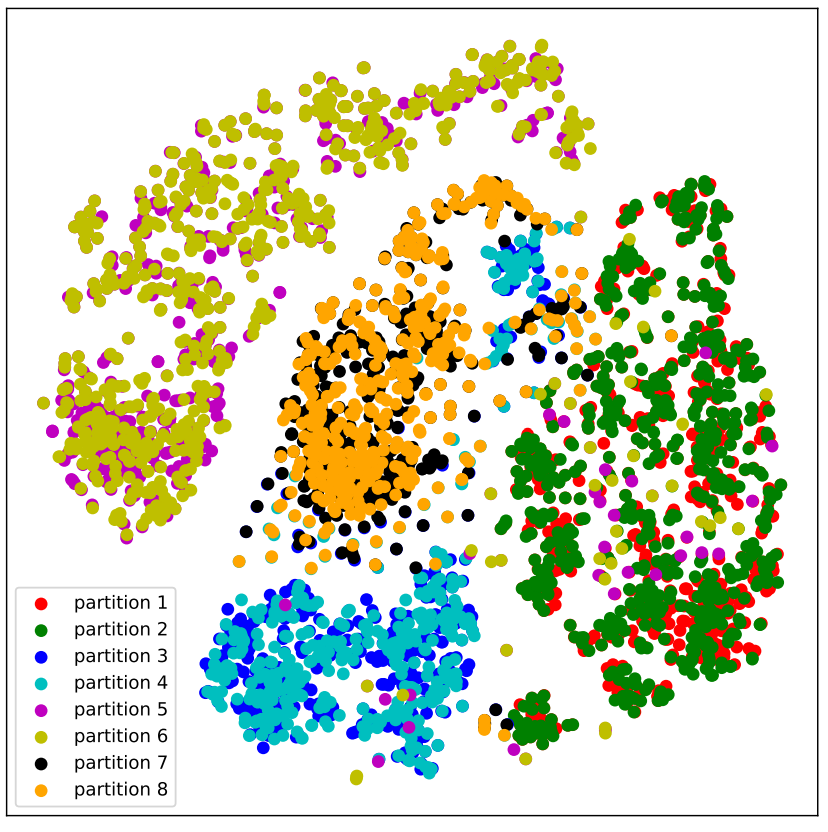}
    \caption{t-SNE visualization}

    \label{fig2}
\end{figure*}
\begin{figure*}[h]
    \centering
    \includegraphics[width=\textwidth]{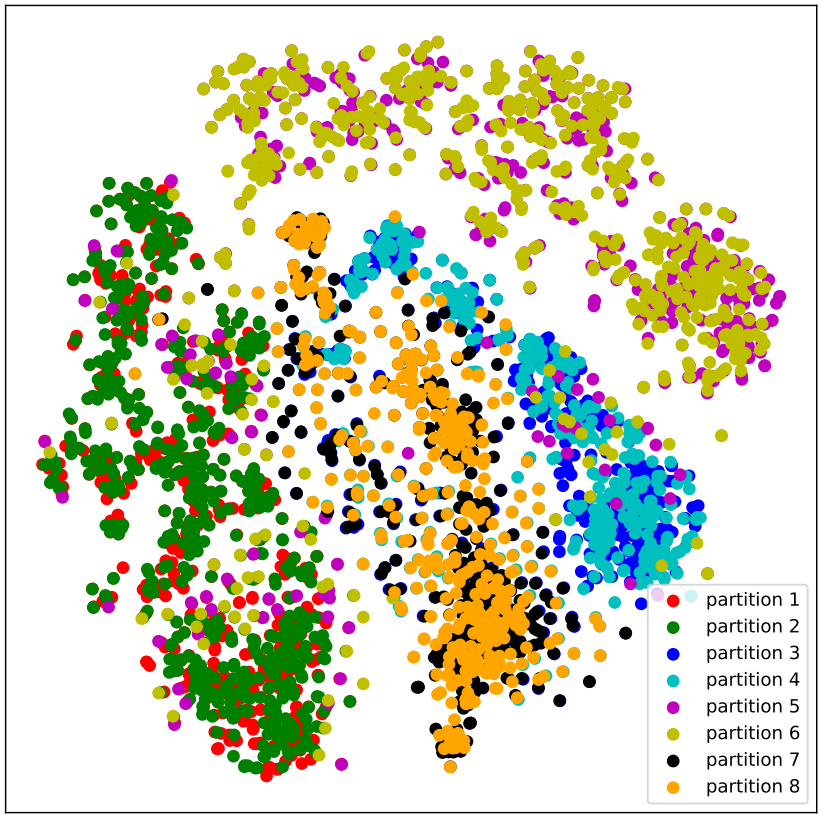}
    \caption{t-SNE visualization}

    \label{fig3}
\end{figure*}
\begin{figure*}[h]
    \centering
    \includegraphics[width=\textwidth]{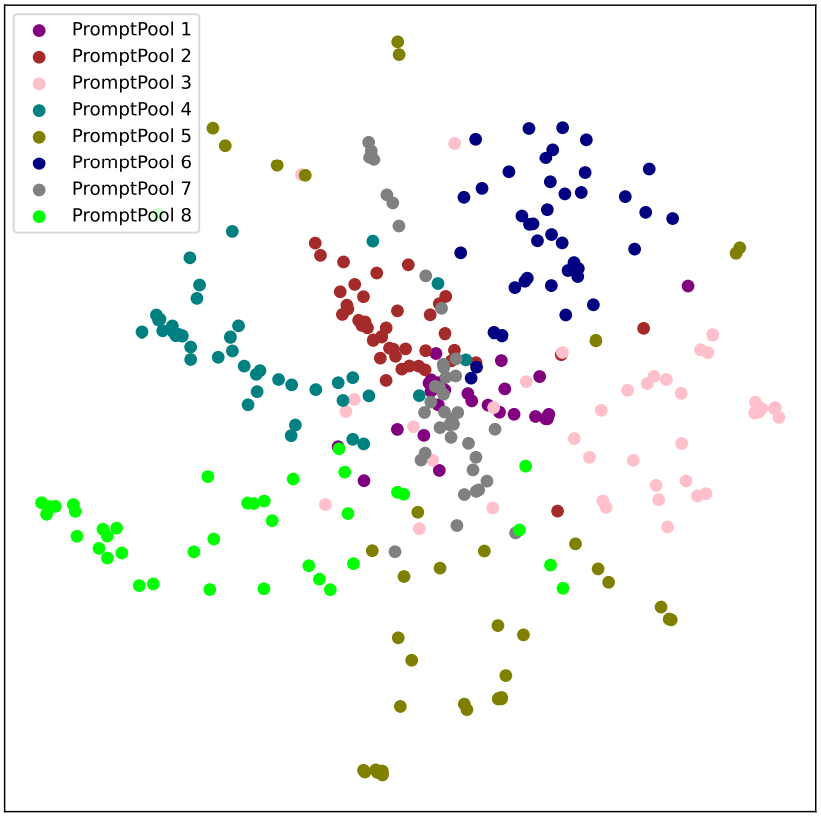}
    \caption{t-SNE visualization}

    \label{fig4}
\end{figure*}
\begin{figure*}[h]
    \centering
    \includegraphics[width=\textwidth]{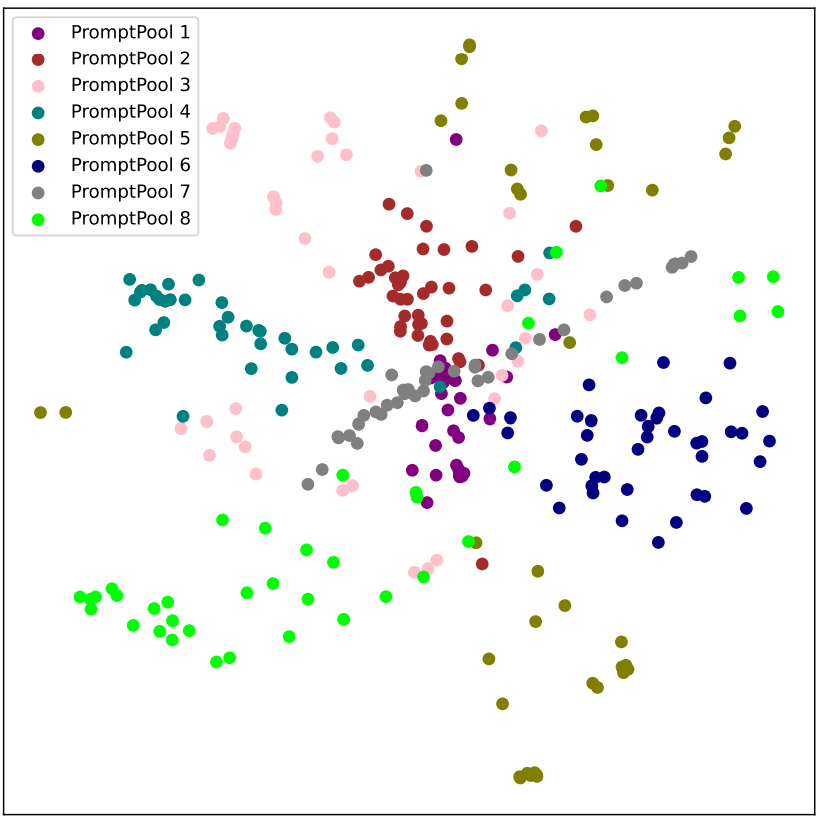}
    \caption{t-SNE visualization}

    \label{fig5}
\end{figure*}
\begin{figure*}[h]
    \centering
    \includegraphics[width=\textwidth]{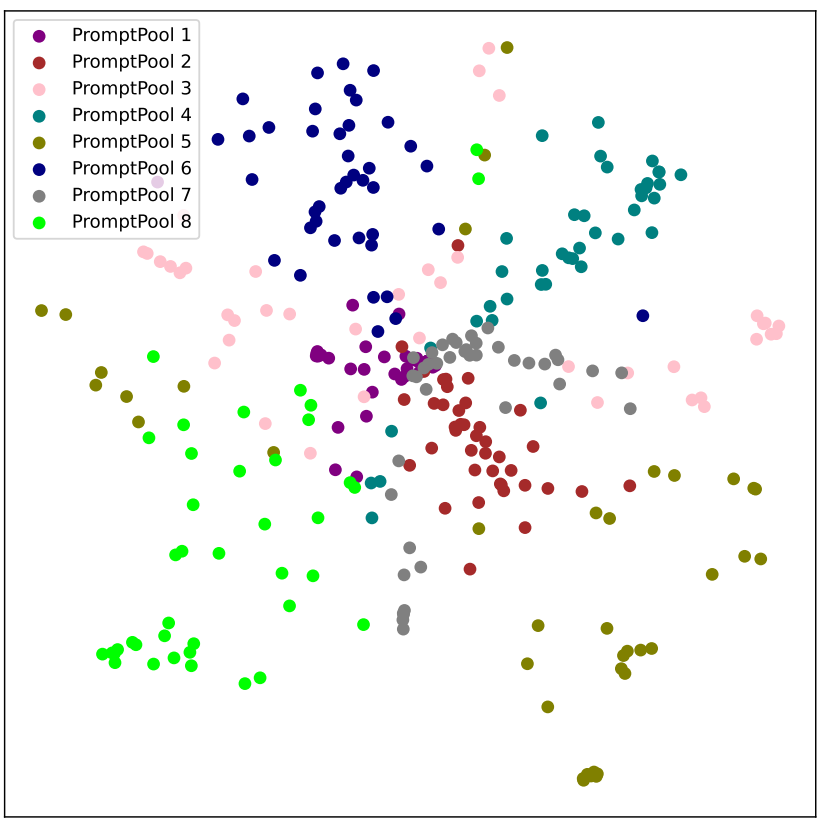}
    \caption{t-SNE visualization}

    \label{fig6}
\end{figure*}